\documentclass[conference]{IEEEtran}
\IEEEoverridecommandlockouts
\usepackage{subfig}
\usepackage{tablefootnote}
\usepackage{multirow}
\usepackage{amsmath,amssymb,amsfonts}
\usepackage{algorithmic}
\usepackage{graphicx}
\usepackage{textcomp}
\usepackage{xcolor}
\usepackage{todonotes}
\usepackage[english]{babel}
\usepackage{hyperref}
\addto\extrasenglish{

}

\usepackage[
    backend=biber,
    style=ieee,
  ]{biblatex}
\newcommand\diff[1]{\textcolor{black}{#1}}
\newcommand\dif[1]{\textcolor{black}{#1}}
\newcommand\di[1]{\textcolor{black}{#1}}

\def\BibTeX{{\rm B\kern-.05em{\sc i\kern-.025em b}\kern-.08em
    T\kern-.1667em\lower.7ex\hbox{E}\kern-.125emX}}
\begin{document}

\title{Irrelevant Pixels are Everywhere: Find and Exclude Them for More Efficient Computer Vision\\
\thanks{This project is supported in part by NSF OAC-2104709 and NSF OAC-2107020. Any opinions, findings, conclusions, or recommendations expressed in this material by the authors do not necessarily reflect the views of the sponsors.}
}

\author{\IEEEauthorblockN{Caleb Tung\textsuperscript{1}, Abhinav Goel\textsuperscript{1}, Xiao Hu\textsuperscript{1}, Nick Eliopoulos\textsuperscript{1}, Emmanuel S. Amobi\textsuperscript{2},
\\George K. Thiruvathukal\textsuperscript{2}, Vipin Chaudhary\textsuperscript{3}, Yung-Hsiang Lu\textsuperscript{1}}
\IEEEauthorblockA{\textsuperscript{1}Purdue University, \textsuperscript{2}Loyola University Chicago, \textsuperscript{3}Case Western Reserve University\\
\{tung3, goel39, hu440, neliopou, yunglu\}@purdue.edu, eamobi@luc.edu, gkt@cs.luc.edu, vxc204@case.edu}
}

\maketitle

\begin{abstract}
Computer vision is often performed using Convolutional Neural Networks (CNNs).
CNNs are compute-intensive and challenging to deploy on power-contrained systems such as mobile and Internet-of-Things (IoT) devices.
CNNs are 
compute-intensive because they indiscriminately compute \diff{many} features on all pixels of the input image.
We observe that, given a computer vision task, images often contain pixels that are irrelevant to the task.
For example, if the task is looking for cars, pixels in the sky are not very useful.
Therefore, we propose that a CNN be modified to only operate on relevant pixels to save computation and energy.
We propose a method to study three popular computer vision datasets, finding that 48\% of pixels are irrelevant.
We also propose the focused convolution to modify a CNN's convolutional layers to reject the pixels that are marked irrelevant.
On an embedded device, we observe no loss in accuracy, while inference latency, energy consumption, and multiply-add count are all reduced by about 45\%.

\end{abstract}


\section{Introduction}
\label{sec:introduction}

Convolutional Neural Networks (CNNs) are known for their high accuracy at many computer vision tasks. However, this accuracy comes at a cost: CNNs are compute-intensive. \diff{ResNet, a popular computer vision CNN for performing the relatively simple task of image classification, 
needs to compute nearly 30 million parameters across all the pixels in the input image}~\cite{thiruvathukal_low-power_nodate}. This high compute requirement is typically satisfied by running the CNN on powerful Graphics Processing Units (GPUs) or other hardware accelerators. However, \textit{low-power systems} (i.e., Internet-of-Things (IoT), mobile, and embedded devices) often impose power and memory constraints that make it challenging to deploy CNNs on them~\cite{thiruvathukal_low-power_nodate}.


To lessen the computation of a CNN on low-power systems, many methods opt to reduce the sheer magnitude of CNN parameters. These include \textit{pruning}~\cite{wang_structured_2020} and \textit{quantization}~\cite{courbariaux_binarized_2016} to cut away redundant parameters or reduce precision, and further include \textit{knowledge distillation}~\cite{peng_correlation_2019} and \textit{neural architecture search}~\cite{tan_mnasnet_2019} to train small CNNs with fewer parameters.

\dif{
These methods all improve efficiency by changing the computer vision model itself; we instead propose changing the \textit{input} to the model. 
CNNs operate indiscriminately on every single pixel in the input; therefore, if we reduce the input, we reduce the computation.}

\dif{In this paper, we \di{propose} that input images have many pixels that can be deleted, and that by doing so, a CNN would save energy. We confirm this idea by creating methods to (1)~identify that three different popular computer vision datasets all contain many such pixels and (2)~demonstrate the energy and inference speed improvements possible by using our focused convolution to delete those pixels.}

\begin{figure}[t]
    \centering
    \subfloat[]{
        \includegraphics[width=0.48\linewidth]{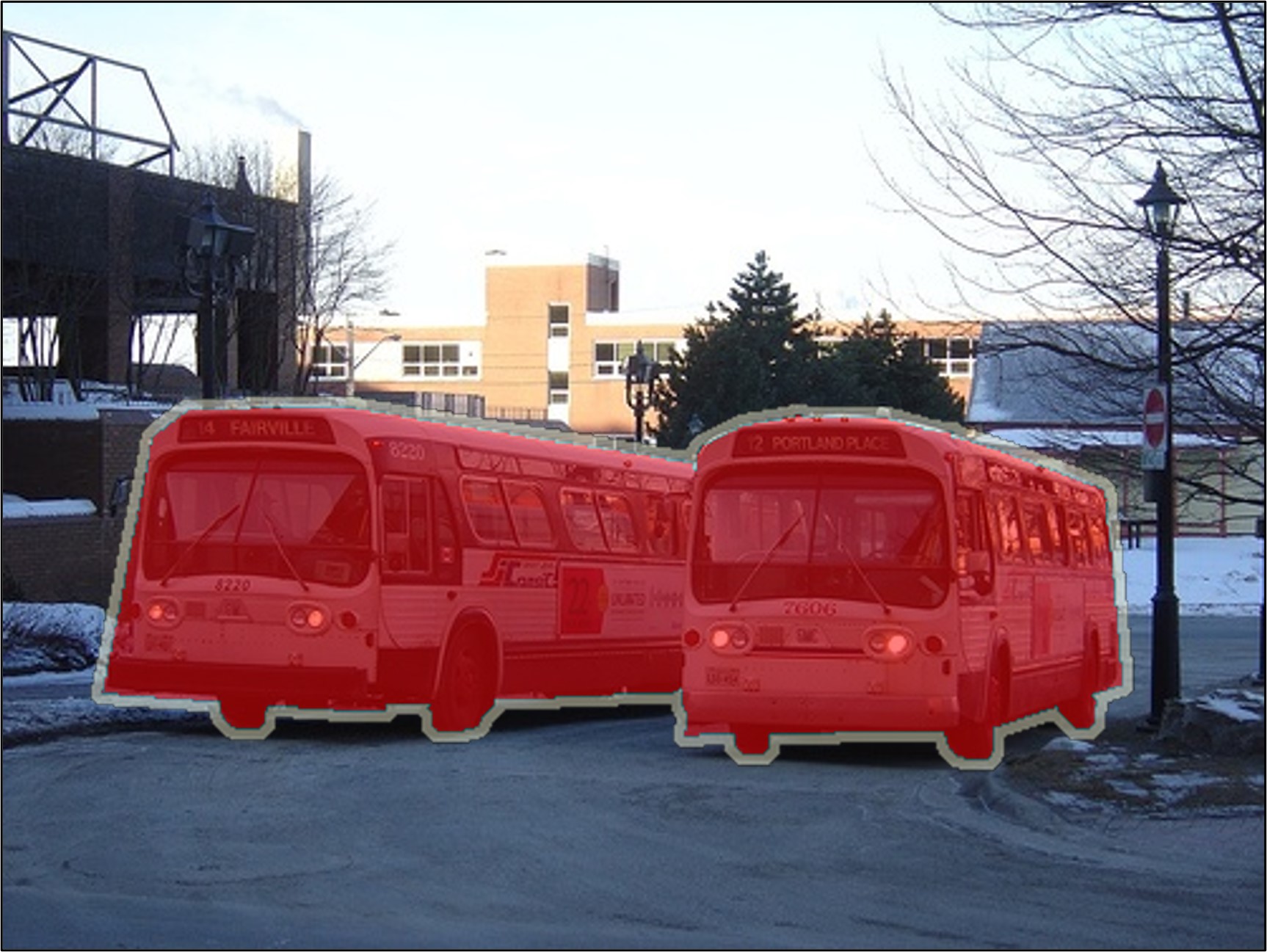}
    }
    \subfloat[]{
        \includegraphics[width=0.48\linewidth]{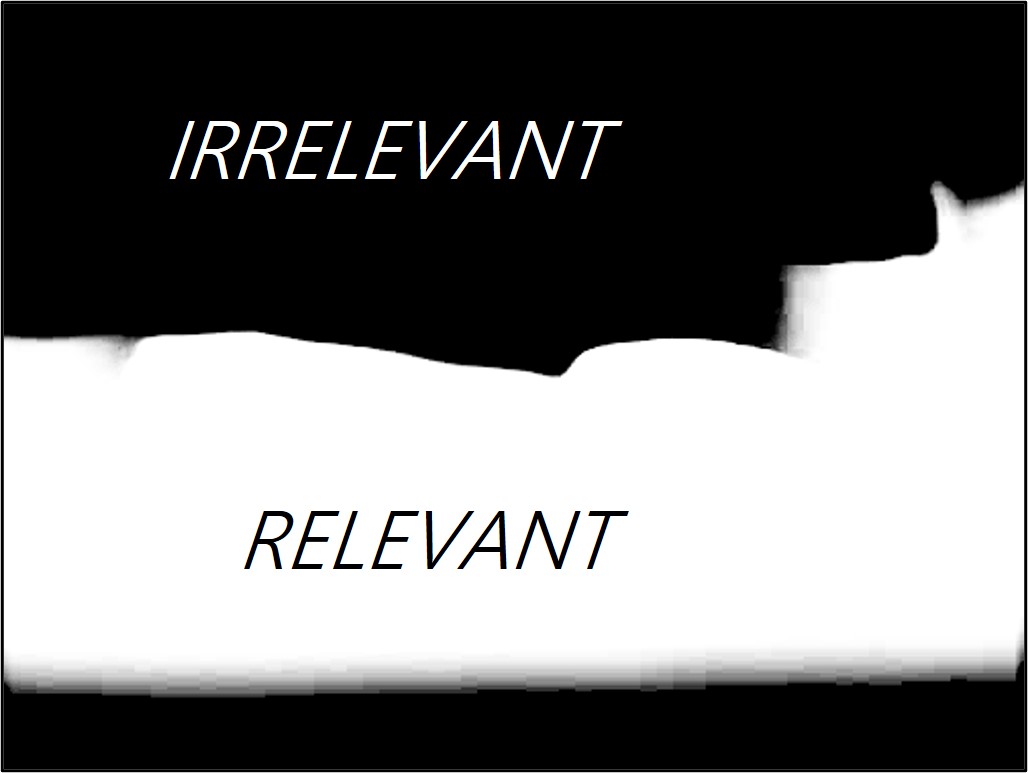}
    }
    
    \caption{\small Example of irrelevant pixels in a PASCAL VOC dataset image. \textbf{(a)} The shuttle buses are highlighted in red. \textbf{(b)} Our method shows that a significant number of pixels (black) are irrelevant, providing little utility, even though CNNs waste computation on them.}
    \label{fig:irrelevant-pixels-example}
    \vspace{-0.1in}
\end{figure}

\dif{An \textit{irrelevant pixel} (\autoref{fig:irrelevant-pixels-example}) (formally defined in \autoref{sec:irrelevant-pixels-in-popular-datasets}) is one that is not useful for the computer vision task (e.g., a building's pixels are not useful when looking for cars). We use depth maps to convert ground truth labels into irrelevant pixel maps, showing that in the Microsoft Common Objects in Context (COCO, 164,000+ images)~\cite{lin_microsoft_2014}, Multi-Object Tracking Challenge (MOT Challenge, 5,500 images)~\cite{leal-taixe_motchallenge_2015}, and PASCAL VOC~\cite{everingham_pascal_2010} (17,900+ images) datasets, roughly 48\% of pixels are irrelevant.}

\dif{Given the irrelevant pixels, we further experiment to find that explicitly excluding those pixels during inference significantly reduces a CNN's compute expenses, saving energy and speeding inference. Our proposed \textit{focused convolution} technique modifies a CNN such that the model itself can exclude pixels marked as irrelevant while still using the same parameters and General Matrix-Multiplication convolutional technique. Similarly inspired work includes Uber's SBNet and its variations; SBNet does convolution in ResNet-sized blocks and tries to fit the blocks into the relevant pixels~\cite{ren_sbnet_2018, verelst_dynamic_2020}. Those other methods still require the model to be changed and re-trained, ours does not. Replacing normal convolution with the focused convolution on the three datasets reduces multiply-add operations, energy consumption, and inference latency by about 45\% in two popular CNNs.}

This paper's contributions: (1) use depth maps to find how many pixels are irrelevant in three popular computer vision datasets, and (2) \di{demonstrate} that CNNs can save on inference latency and energy consumption by excluding irrelevant pixels from computation and only performing operations on relevant ones, via our focused convolution.
\section{\dif{Finding Irrelevant Pixels in Datasets}}
\label{sec:irrelevant-pixels-in-popular-datasets}
We propose a method to study datasets for irrelevant pixels. We find that irrelevant pixels are commonly found in the COCO, PASCAL VOC, and MOTChallenge datasets.

\subsection{Counting Irrelevant Pixels with Depth Maps}
\label{subsec:using-depth-maps-to-count-irrelevant-pixels}
\dif{Our study defines a dataset's relevant pixels as: \textit{all pixels that comprise the dataset ground truth objects and their associated depth levels.} In \autoref{fig:irrelevant-pixels-example}a, the ground truth pixels are represented by the red area around the shuttle buses. The final pixelwise map of relevant pixels, shown in \autoref{fig:irrelevant-pixels-example}b, is generated using depth level thresholding (explained below) and includes additional pixels.}

\begin{figure}
    \centering
    \subfloat[]{
        \includegraphics[width=0.47\linewidth]{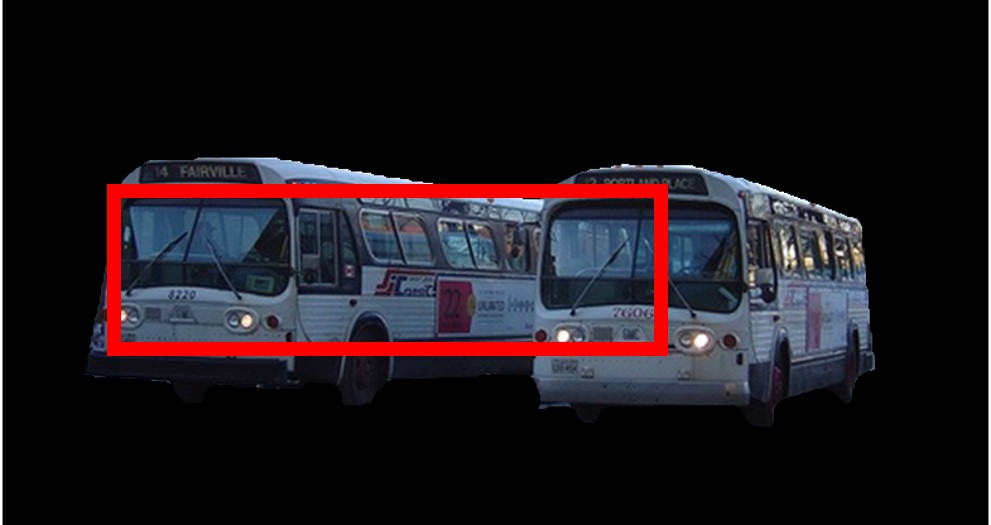}
    }
    \subfloat[]{
        \includegraphics[width=0.47\linewidth]{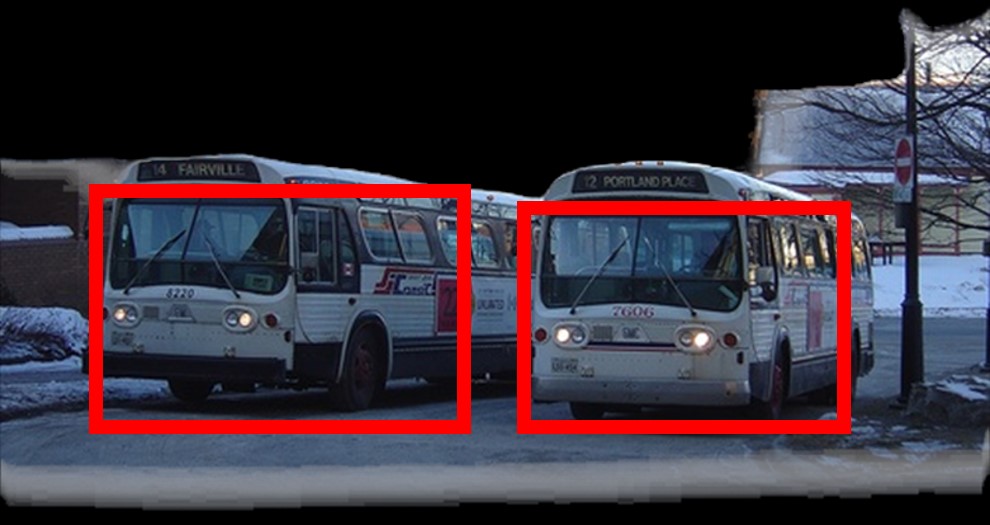}
    }
    \caption{\small Ground truth pixels alone \textbf{(a)} are not always a sufficient amount of relevant pixels for a CNN to make good predictions (red box is inaccurate, misses second bus); contextual pixels at similar depth levels are also needed \textbf{(b)} for CNN accuracy (two good detections).}
    \label{fig:ground_truth_only}
    \vspace{-0.1in}
\end{figure}

\begin{figure}
    \centering
    \subfloat[]{
        \includegraphics[width=0.9\linewidth]{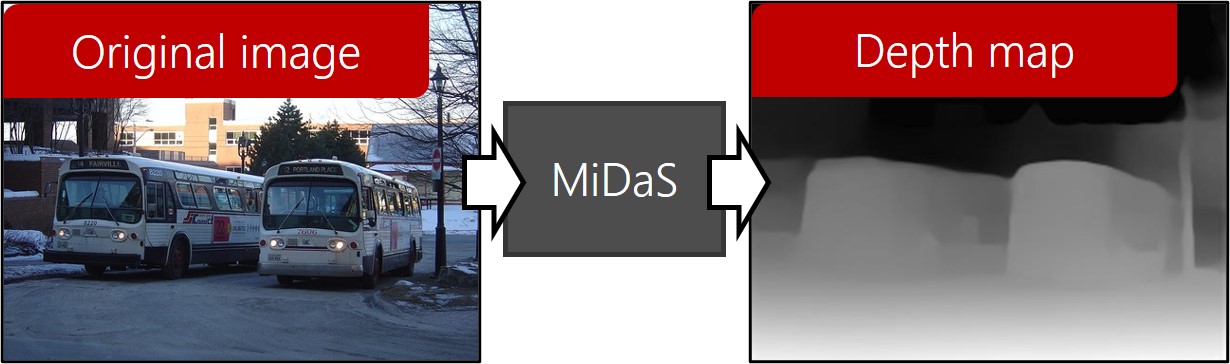}
    }
    
    \subfloat[]{
        \includegraphics[width=0.9\linewidth]{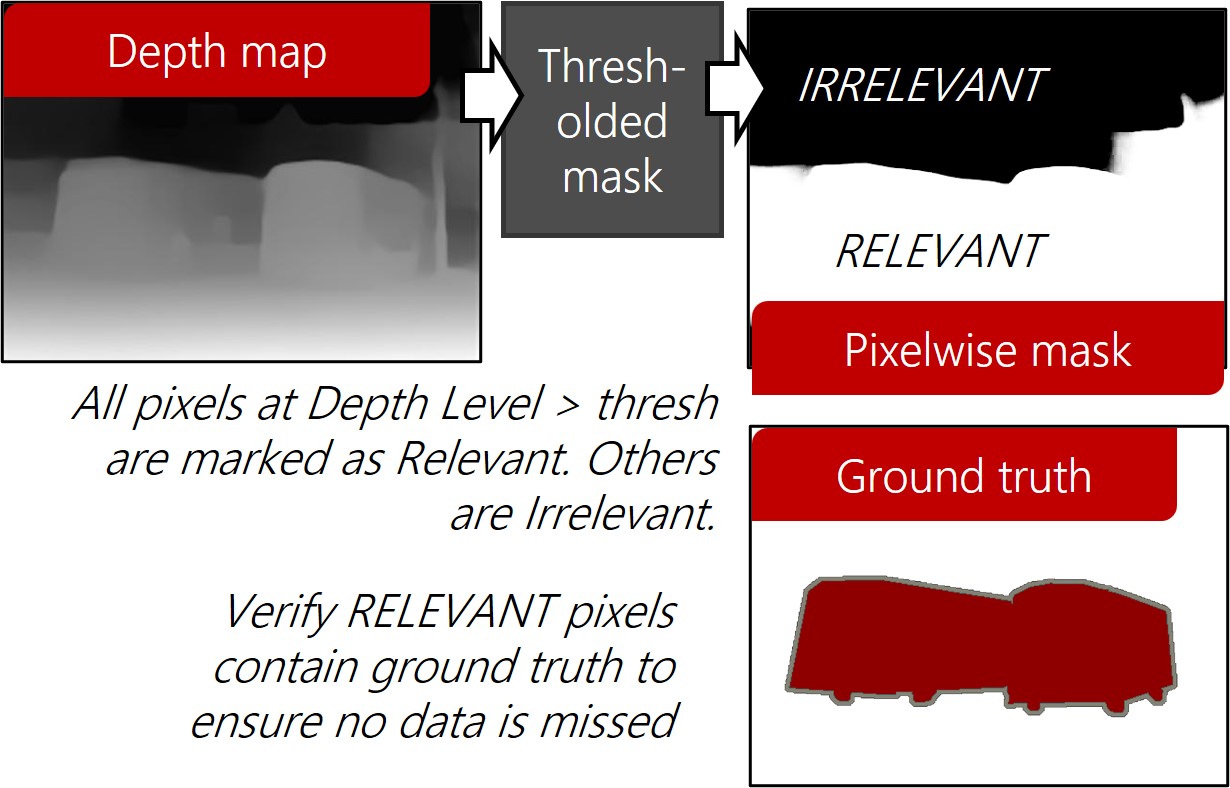}
    }
    
    \caption{\small This is how to count the number of irrelevant pixels in datasets: First, use MiDaS to generate the image's depth map \textbf{(a)}. Next, threshold the depth map, producing a small Relevant Pixels region \textbf{(b)}. Verify against ground truth to ensure all important data is captured.}
    \label{fig:counting_irrelevant_pixels}
\end{figure}

\dif{Ground truth pixels alone do not represent all relevant pixels. A typical CNN uses not only the pixels comprising the object (\autoref{fig:ground_truth_only}a), but also the \textit{pixels from the surrounding area of an object} (\autoref{fig:ground_truth_only}b) to extract features that contextualize and understand the object~\cite{gauen_comparison_2017}. To properly include these contextual pixels, our method uses \textit{depth maps}. Shown in \autoref{fig:counting_irrelevant_pixels}a, an image's depth map represents each pixel in the image with one value, referred to as the pixel's \textit{depth level}, denoting how far away from the camera that pixel is.}



\dif{Pixels that have similar depth levels to a given object on the ground are known to provide useful, contextual pixels that surround the object~\cite{song_semantic_2017}. For example, a car will be at a similar depth level as its context: the road on which it drives. Therefore, by thresholding the depth level appropriately, we can capture all the relevant pixels: both the contextual pixels and their associated objects.}

\dif{We generate depth maps with ``MiDaS,'' a depth estimation model by Ranftl, et al.~\cite{ranftl_towards_2020} MiDaS is chosen from among other monocular depth estimation techniques because it is uniquely trained using a mix of diverse datasets, dramatically improving its performance on Microsoft COCO, our largest target dataset.}

Our method thresholds the depth map to create a binary, pixelwise mask of relevant and irrelevant pixels, such that the relevant pixel region encapsulates the ground truth (see \autoref{fig:counting_irrelevant_pixels}). Using the observation that camera focal lengths tend to cause pictures to be taken at mid-range depth levels~\cite{gauen_comparison_2017}, we threshold an image's MiDaS depth map according to the mid-range of its distribution, filtering out pixels at short-range and long-range depth levels.  The filtered pixels are considered irrelevant and the remaining pixels are considered relevant. We verify this binary, pixelwise mask against the ground truth labels: if the mask's relevant-marked region contain the ground truth, then we assume all ground truth and associated contextual pixels are captured, and we record the number of irrelevant VS relevant pixels. If not, the threshold is expanded and the process is repeated.

\begin{figure}[t]
    \centering
    \includegraphics[width=0.9\linewidth]{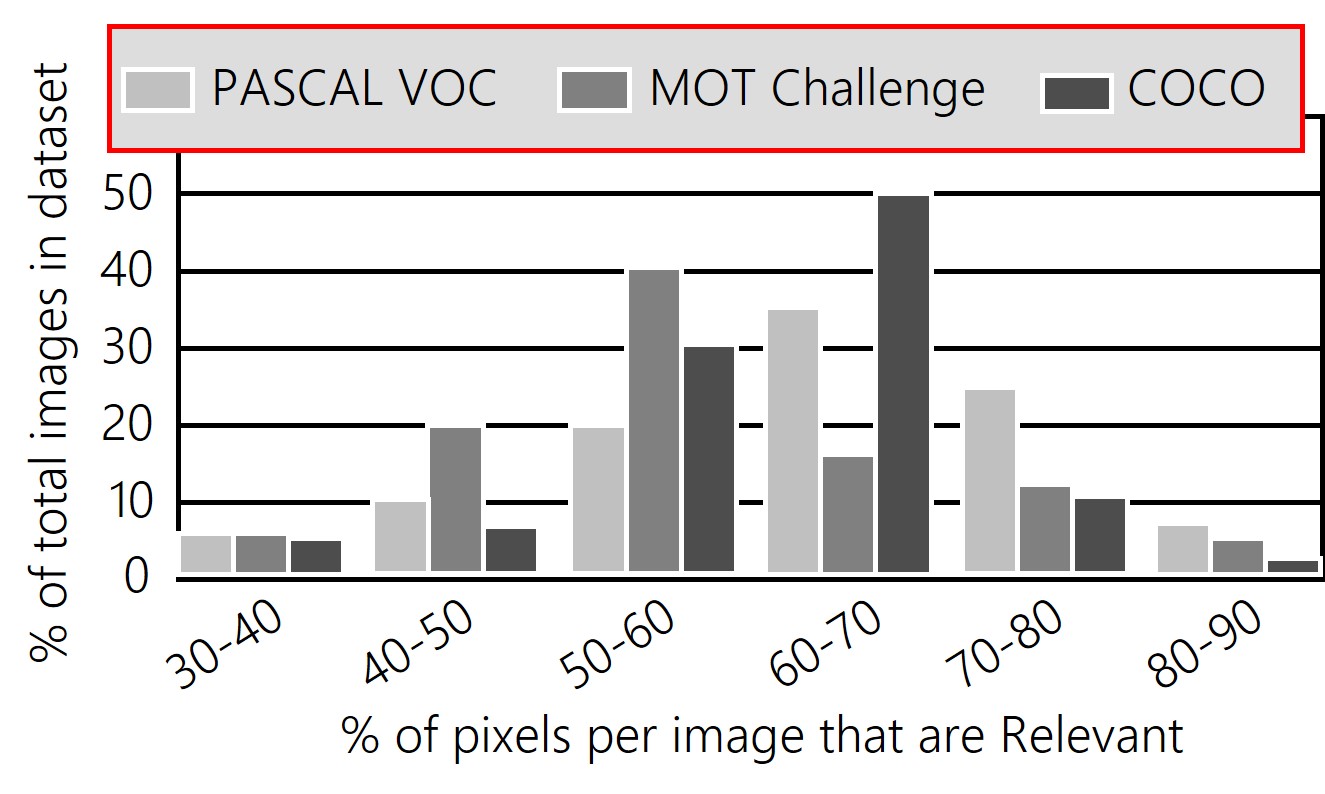}
    \caption{ \small Average percentage of pixels that are relevant. Less than 1\% of images contain 10-30\% or 90-100\% useful pixels and are not shown on the graph. As shown, most images contain 50\%-60\% relevant (40-50\% irrelevant pixels). This means that in most images, 40-50\% of pixels do not contribute to useful CNN computations and, therefore, can be excluded from processing.}
    \label{fig:distribution-irrelevant-pixels}
    \vspace{-0.1in}
\end{figure}

\subsection{Exploring Datasets for Irrelevant Pixels}
\label{subsec:exploring_datasets_for_irrelevants_pixels}
In our study, we use our depth map method to count irrelevant pixels in each image in the three datasets, totaling nearly 200,000 images. Results are shown in \autoref{fig:distribution-irrelevant-pixels}. Irrelevant pixels are quite common in the three datasets. On average, 42\% of all pixels are irrelevant in PASCAL VOC, 49\% are irrelevant in MOT Challenge, and 45\% are irrelevant in COCO. This suggests that CNNs are wasting significant amounts of compute resources on embedded devices by computing convolutions on all those irrelevant pixels.

\section{\dif{Excluding Irrelevant Pixels in Datasets}}



To demonstrate that being able to exclude irrelevant pixels can significantly save computer vision energy consumption and inference latency, we propose the \textit{focused convolution} technique.
In typical CNN computer vision models, the 2D-convolutional layer is responsible for up to 80\% of the computations performed by the neural network~\cite{thiruvathukal_low-power_nodate}. Standard convolution layers are not designed to exclude irrelevant pixels. Instead, they operate on all the input pixels, wasting computation on irrelevant pixels. Because CNNs have multiple convolutional layers in succession, this waste is repeated across multiple layers, compounding the negative impact on the model's energy use and speed. The focused convolution reduces this waste by excluding any pixels marked irrelevant \dif{by some assumed oracle - in this case, our relevant VS irrelevant pixelwise masks generated in \autoref{subsec:exploring_datasets_for_irrelevants_pixels}.}

\subsection{Using Focused Convolutions to Exclude Irrelevant Pixels}
\label{subsec:using_focused_convolutions_to_exclude_irrelevant_pixels}
\begin{figure}[t]
    \centering
    \subfloat[]{
        \includegraphics[width=0.9\linewidth]{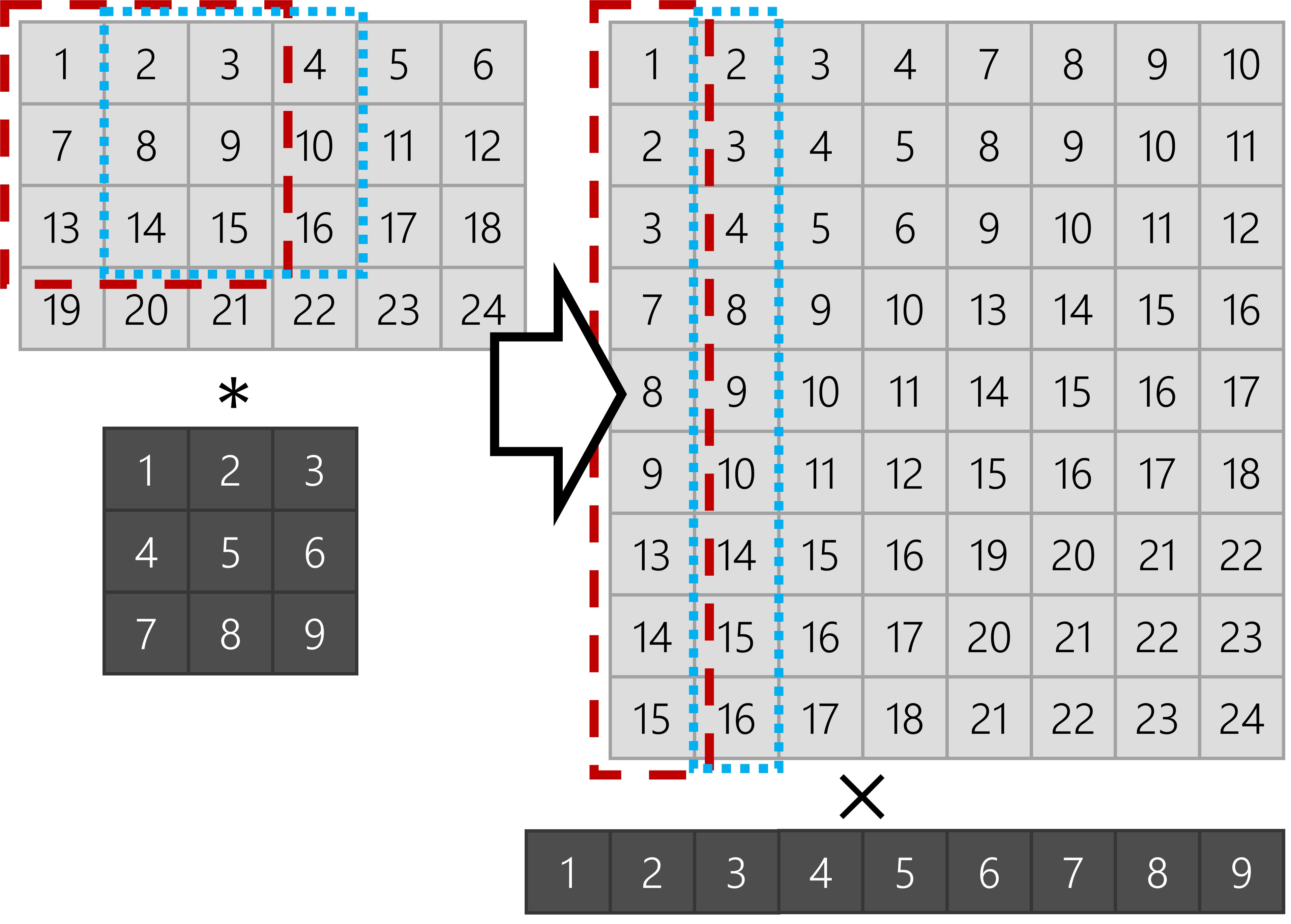}
        \label{subfig:gemm_explained}
    }
    
    \subfloat[]{
        \includegraphics[width=0.9\linewidth]{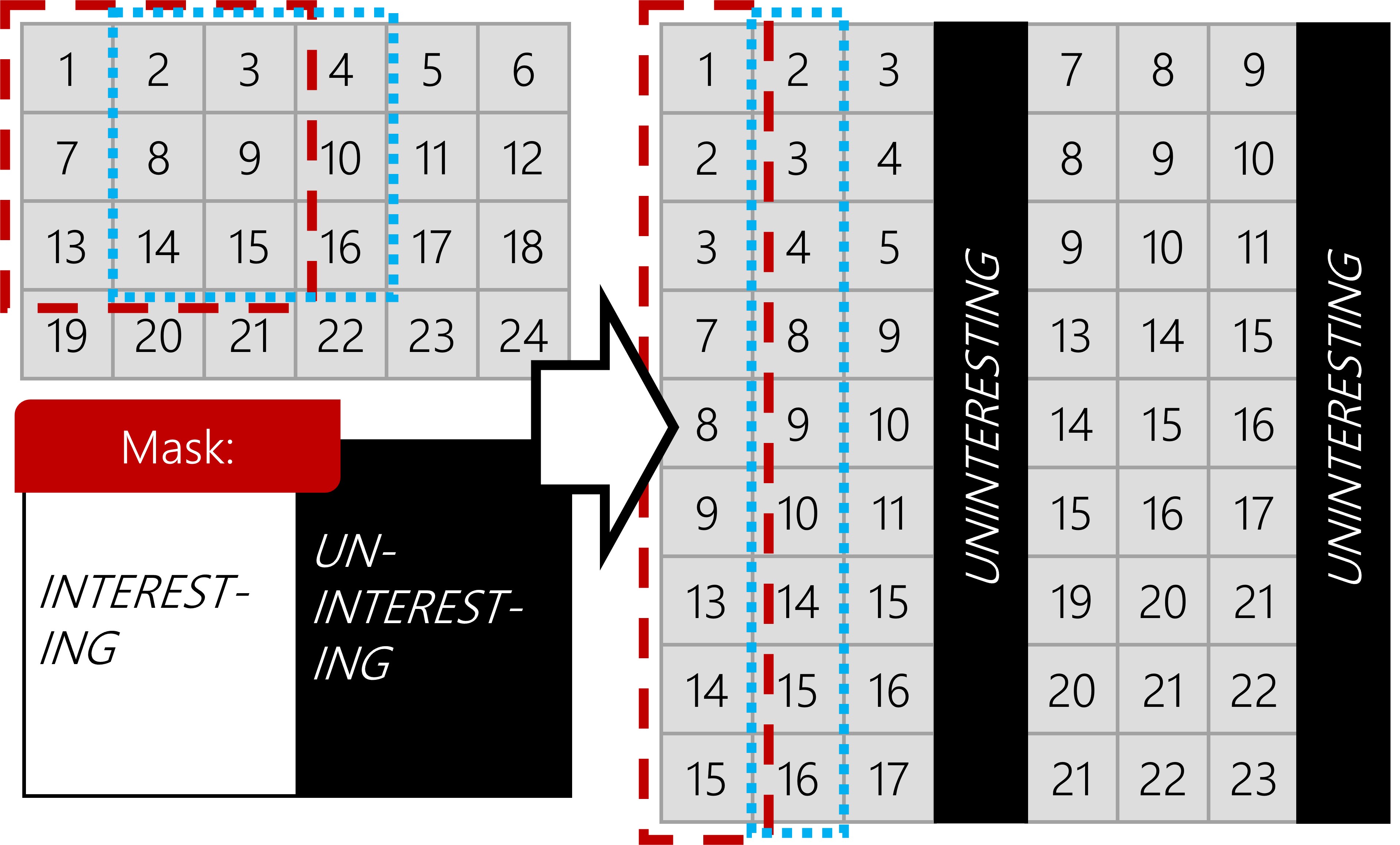}
        \label{subfig:focused_convolution_explained}
    }
    
    \caption{\small \textbf{(a)} Traditional GEMM convolution uses the \textit{im2col} procedure to convert a $1 \times 4 \times 6 \times 1$ input tensor (light gray), assuming a stride-1, $3 \times 3$ weight kernel (dark gray), to a $9 \times 8$ matrix for matrix multiplication. Each $9\times 1$ column of the matrix (e.g., red, blue dashed line) represents one $3 \times 3$ patch from the input tensor. \textbf{(b)} The proposed \textit{focused convolution} modifies \textit{im2col} exclude patches deemed irrelevant by an oracle-generated pixelwise mask. Thus, the final $9\times 6$ matrix will have fewer columns than that of traditional GEMM, resulting in less computation on embedded devices.}
    
\end{figure}

Our focused convolution \di{improves} the popular General Matrix Multiplication (GEMM) technique~\cite{anderson_low-memory_2017}. The GEMM technique reduces a sequential, sliding-window 2D-convolution to a parallelized, matrix-multiplication operation; matrix multiplications are heavily optimized on modern computers~\cite{anderson_low-memory_2017}. GEMM (Figure~\autoref{subfig:gemm_explained}) uses a procedure called \textit{im2col} to convert patches of the 3D input tensor (e.g., an RGB image) into the columns of a 2D matrix, and then multiplies the matrix by the convolutional weights to retrieve the results of convolution.

The focused convolution \di{enables} the \textit{im2col} procedure to accommodate a pixelwise mask (such as those generated in \autoref{subsec:exploring_datasets_for_irrelevants_pixels}) indicating whether a pixel is relevant or not. If a patch of pixels is labeled irrelevant, then the modified \textit{im2col} will not convert the patch into a column of the matrix. Thus, those irrelevant pixels get excluded entirely by the matrix multiplication, as shown in Figure\autoref{subfig:focused_convolution_explained}.

\dif{Because the focused convolution is simply designed to accommodate the pixelwise masks as inputs, it does not need to change the model itself.} It can be used in any pretrained CNN, requiring no additional training. It can replace existing convolution layers without changing the weights and biases. If no pixelwise mask is supplied, the focused convolution behaves identically to a standard convolution.

For a given image and convolutional layer, we can count the number of operations needed to complete a GEMM convolution's matrix multiplication. Assume the input has dimensions $B_I \times C_I \times H_I \times W_I$ (batch, channels, height, width), and the convolutional weights has dimensions $S_W \times S_W$ (side lengths) with stride $s$ and no padding. A normal \textit{im2col} would convert each channel of the input to $\frac{H_I-S_W}{s}\frac{W_I-S_W}{s}$ columns of $S_WS_W$ pixels. That amounts to multiplying the weights with a $S_WS_W \times C_I\frac{(H_I-S_W)}{s}\frac{W_I-S_W}{s}$ matrix, $B_I$ times. That is a total of $B_I\frac{C_I(H_I-S_W)}{s}\frac{W_I-S_W}{s}S_WS_W$ multiply/add operations.

If we convert the GEMM convolution to a focused convolution, we save matrix multiplication operations by excluding columns of irrelevant pixels. As we discovered in \autoref{subsec:exploring_datasets_for_irrelevants_pixels}, an average of 48\% of pixels are irrelevant in popular datasets. If the region of relevant pixels is a rectangle that takes up $p$\% of the pixels, then the focused convolution would produce only $p\%\frac{H_I-S_W}{s}\frac{W_I-S_W}{s}$ columns, resulting in a $100-p$\% reduction in operations. Therefore, the focused convolution's energy and latency improvements is a function of the linear relationship between number of irrelevant pixels and image size.


\subsection{Evaluating the Compute Savings of Focused Convolutions}
\label{subsec:evaluating_the_compute_savings_of_focused_convolutions}
We implement and test the focused convolution using PyTorch on a Raspberry Pi 3 (average power 5 W). We compare the focused convolution with a normal PyTorch GEMM convolution. Excluding pixels is beneficial on more powerful hardware, too - we compare the focused convolution with an Intel MKL-optimized convolution~\cite{wang_intel_2014} on a Intel Core i7 CPU (average power 28 W). For input images, we use the MOT Challenge, COCO, and PASCAL VOC. We exclude the pixels deemed irrelevant by our method from \autoref{subsec:using-depth-maps-to-count-irrelevant-pixels}.

Our tests comprise of two popular computer vision CNNs designed for use on low-power devices: EfficientDet~\cite{tan_efficientdet_2020} and SSD-Lite~\cite{sandler_mobilenetv2_2018}. We use PyTorch's pretrained weights, and we replace each CNN's convolutional layer with a focused convolutional layer. The pixelwise, binary Relevant-VS-Irrelevant masks we generated in \autoref{subsec:exploring_datasets_for_irrelevants_pixels} are  propagated through the CNN by scaling the mask's size for each  layer.

\autoref{tab:focused_conv_results} summarizes the observed energy consumption/inference latency improvements from the focused convolutions. As shown, we see that equipping a CNN with focused convolution allows it to dramatically reduce its computation expense by excluding irrelevant pixels on images from popular datasets. On Intel CPU, the MKL optimizations allow a fullsize, ``wasteful'' convolution to operate comparably to the focused convolution, but such optimizations are unavailable on low-power devices, so the focused convolution is preferable.

Finally, we verify the focused convolution-equipped CNN's detection accuracy remains identical to the original CNN.

\subsection{Assumptions, Challenges, and Opportunities}
This paper assumes that including all pixels at the same depth level as ground truth accurately captures all related pixels. This may not be true when a camera's focus is extended (e.g. wide, scenic shots). Still, our technique is demonstrably sufficient for three major datasets. Additionally, depth mapping is computationally intensive, so future irrelevant pixel generation will need to either be offloaded or rapidly generated.


\begin{table}[t]
\begin{tabular}{|rlrrrrrr|}
\hline
\multicolumn{2}{|l|}{\multirow{2}{*}{}}                                                                                 & \multicolumn{2}{l|}{\textit{MOT2015}}                                     & \multicolumn{2}{l|}{\textit{COCO}}                                        & \multicolumn{2}{l|}{\textit{PASCAL VOC}}                               \\ \cline{3-8} 
\multicolumn{2}{|l|}{}                                                                                                  & \multicolumn{1}{l|}{\textit{ED}}    & \multicolumn{1}{l|}{\textit{SL}}    & \multicolumn{1}{l|}{\textit{ED}}    & \multicolumn{1}{l|}{\textit{SL}}    & \multicolumn{1}{l|}{\textit{ED}}    & \multicolumn{1}{l|}{\textit{SL}} \\ \hline
\multicolumn{8}{|l|}{\textit{Number of Mult-Add Operations (M/inference)}}                                                                                                                                                                                                                                                                                    \\ \hline
\multicolumn{2}{|r|}{Normal}                                                                                            & \multicolumn{1}{r|}{384.5}          & \multicolumn{1}{r|}{483.6}          & \multicolumn{1}{r|}{384.5}          & \multicolumn{1}{r|}{483.6}          & \multicolumn{1}{r|}{384.5}          & 483.6                            \\ \hline
\multicolumn{2}{|r|}{\textbf{Focused}}                                                                                  & \multicolumn{1}{r|}{\textbf{196.1}} & \multicolumn{1}{r|}{\textbf{246.8}} & \multicolumn{1}{r|}{\textbf{211.4}} & \multicolumn{1}{r|}{\textbf{266.0}} & \multicolumn{1}{r|}{\textbf{223.0}} & \textbf{280.4}                   \\ \hline
\multicolumn{8}{|l|}{\textit{Inference Latency (s/inference)}}                                                                                                                                                                                                                                                                                           \\ \hline
\multirow{2}{*}{\textit{\begin{tabular}[c]{@{}r@{}}RPi\\ (5W)\end{tabular}}}    & \multicolumn{1}{l|}{Normal}           & \multicolumn{1}{r|}{2.10}           & \multicolumn{1}{r|}{2.26}           & \multicolumn{1}{r|}{2.00}           & \multicolumn{1}{r|}{2.33}           & \multicolumn{1}{r|}{2.06}           & 2.29                             \\ \cline{2-8} 
                                                                                & \multicolumn{1}{l|}{\textbf{Focused}} & \multicolumn{1}{r|}{\textbf{1.11}}  & \multicolumn{1}{r|}{\textbf{1.30}}  & \multicolumn{1}{r|}{\textbf{1.33}}  & \multicolumn{1}{r|}{\textbf{1.51}}  & \multicolumn{1}{r|}{\textbf{1.47}}  & \textbf{1.56}                    \\ \hline
\multirow{3}{*}{\textit{\begin{tabular}[c]{@{}r@{}}Intel\\ (28W)\end{tabular}}} & \multicolumn{1}{l|}{Normal}           & \multicolumn{1}{r|}{0.25}           & \multicolumn{1}{r|}{0.28}           & \multicolumn{1}{r|}{0.25}           & \multicolumn{1}{r|}{0.29}           & \multicolumn{1}{r|}{0.25}           & 0.28                             \\ \cline{2-8} 
                                                                                & \multicolumn{1}{l|}{MKL}              & \multicolumn{1}{r|}{0.18}           & \multicolumn{1}{r|}{0.19}           & \multicolumn{1}{r|}{0.18}           & \multicolumn{1}{r|}{0.20}           & \multicolumn{1}{r|}{0.18}           & 0.20                             \\ \cline{2-8} 
                                                                                & \multicolumn{1}{l|}{\textbf{Focused}} & \multicolumn{1}{r|}{\textbf{0.17}}  & \multicolumn{1}{r|}{\textbf{0.18}}  & \multicolumn{1}{r|}{\textbf{0.18}}  & \multicolumn{1}{r|}{\textbf{0.20}}  & \multicolumn{1}{r|}{\textbf{0.19}}  & \textbf{0.20}                    \\ \hline
\multicolumn{8}{|l|}{\textit{Energy Consumption (J/inference)}}                                                                                                                                                                                                                                                                                          \\ \hline
\multirow{2}{*}{\textit{\begin{tabular}[c]{@{}r@{}}RPi\\ (5W)\end{tabular}}}    & \multicolumn{1}{l|}{Normal}           & \multicolumn{1}{r|}{10.22}          & \multicolumn{1}{r|}{11.80}          & \multicolumn{1}{r|}{10.15}          & \multicolumn{1}{r|}{11.81}          & \multicolumn{1}{r|}{10.20}          & 10.90                            \\ \cline{2-8} 
                                                                                & \multicolumn{1}{l|}{\textbf{Focused}} & \multicolumn{1}{r|}{\textbf{5.60}}  & \multicolumn{1}{r|}{\textbf{6.11}}  & \multicolumn{1}{r|}{\textbf{6.71}}  & \multicolumn{1}{r|}{\textbf{7.50}}  & \multicolumn{1}{r|}{\textbf{7.44}}  & \textbf{7.80}                    \\ \hline
\multirow{3}{*}{\textit{\begin{tabular}[c]{@{}r@{}}Intel\\ (28W)\end{tabular}}} & \multicolumn{1}{l|}{Normal}           & \multicolumn{1}{r|}{6.61}           & \multicolumn{1}{r|}{7.39}           & \multicolumn{1}{r|}{6.45}           & \multicolumn{1}{r|}{7.42}           & \multicolumn{1}{r|}{6.69}           & 7.81                             \\ \cline{2-8} 
                                                                                & \multicolumn{1}{l|}{MKL}              & \multicolumn{1}{r|}{5.18}           & \multicolumn{1}{r|}{5.09}           & \multicolumn{1}{r|}{5.09}           & \multicolumn{1}{r|}{5.61}           & \multicolumn{1}{r|}{5.23}           & 5.60                             \\ \cline{2-8} 
                                                                                & \multicolumn{1}{l|}{\textbf{Focused}} & \multicolumn{1}{r|}{\textbf{4.76}}  & \multicolumn{1}{r|}{\textbf{5.04}}  & \multicolumn{1}{r|}{\textbf{5.10}}  & \multicolumn{1}{r|}{\textbf{5.60}}  & \multicolumn{1}{r|}{\textbf{5.29}}  & \textbf{5.62}                    \\ \hline
\end{tabular}
\caption{\small By excluding irrelevant pixels from popular datasets (MOTChallenge, COCO, PASCAL VOC), CNNs equipped with our \textbf{Focused} convolutions outperform those without (Normal). On an Intel CPU, MKL-optimized convolutions are comparable in performance to focused convolutions. \\ \\ \footnotesize{RPi: Raspberry Pi 3, Intel: Intel Core i7 CPU, ED: EfficientDet, SL: SSD-Lite, MKL: using Intel MKL optimized convolution.}}
\label{tab:focused_conv_results}
\end{table}


\section{Conclusion}
\label{sec:conclusion}
We observe that in computer vision, CNNs are compute-heavy because they waste time looking at irrelevant pixels.
We propose a thresholded depth mapping technique to study modern computer vision datasets for irrelevant pixels.
We find that an average of 42\%, 49\%, and 45\% of  pixels per image are irrelevant, for three popular datasets (PASCAL VOC, MOT Challenge, and COCO, respectively).
We further propose significantly reducing a convolutional layer's multiply-add operations, energy consumption, and inference latency with our focused convolution: this modifies the layer's standard GEMM operations to ignore irrelevant pixels.
Ignoring the irrelevant pixels we computed on the three datasets, we find that multiply-add operations are reduced by an average of 48.3\%, energy consumption by 42.7\%, and inference latency by 47.4\% for two popular object detection CNNs on an embedded device.


\AtNextBibliography{\footnotesize}
\printbibliography

\end{document}